\documentclass[conference]{IEEEtran}

\IEEEoverridecommandlockouts

\usepackage{cite}
\usepackage{amsmath,amssymb}
\usepackage{booktabs}
\usepackage{tabularx}
\usepackage{array}
\usepackage{graphicx}
\usepackage{url}
\usepackage{xcolor}
\usepackage{balance}
\usepackage{listings}
\usepackage[hidelinks]{hyperref}
\usepackage{comment}

\graphicspath{{treat_formal_assets/}}
\renewcommand{\arraystretch}{1.06}

\newcommand{\none}{\texttt{NONE}}

\begin{document}

\title{TREAT: Evaluating Access to Formal Knowledge across Equivalent Mathematical Representations}

\author{
\IEEEauthorblockN{Fateme Mazdarani}
\IEEEauthorblockA{
School of Computing\\
Clemson University\\
Clemson, SC, USA\\
fmazdar@clemson.edu
}
\and
\IEEEauthorblockN{Carlos Toxtli}
\IEEEauthorblockA{
School of Computing\\
Clemson University\\
Clemson, SC, USA\\
ctoxtli@clemson.edu
}
}

\maketitle

\begin{abstract}
AI systems increasingly operate between flexible input representations and formal objects used by downstream tools. A key challenge is recognizing when an unfamiliar formulation denotes a known formal object. We study this challenge through theorem recognition: given an equivalence-preserving transformation of a theorem condition, a model must recover the theorem identity associated with the standard statement. 
We introduce TREAT, a benchmark for evaluating whether large language models can recover known theorem identities from equivalence-preserving formula-level transformations. Rather than paraphrasing theorem text, TREAT changes the mathematical form of theorem conditions themselves, expressing known results through residual equations, witness statements, optimization identities, set relations, operator forms, and proof-intermediate characterizations. Starting from scraped theorem pages, we filter for entries with usable mathematical expression forms, extract canonical theorem conditions, and generate transformed variants with recorded assumptions and inverse mappings. The final corpus contains 737 theorem identities and 29,480 transformed rows. On a test panel, the best model retrieves the correct theorem identity in only 60.73\% of cases. Other systems reveal different failure modes, including abstention, wrong detection, and malformed outputs. These suggest that theorem knowledge can be fragile under equivalent changes in representation. TREAT therefore provides a controlled testbed for evaluating representation-robust access to formal knowledge, with broader relevance to domains that require stable target objects, explicit equivalence relations, validation procedures, and auditable scoring.

\end{abstract}

\begin{IEEEkeywords}
language models, representation robustness, formal knowledge retrieval, theorem recognition, mathematical reasoning, benchmark datasets, equivalence-preserving transformations, structured prediction
\end{IEEEkeywords}

\section{Introduction}
A central requirement for intelligent systems is the ability to access relevant knowledge across changes in representation. For AI systems, this creates a representation-access problem: the same underlying concept may appear through many equivalent surface forms, but only some of them may activate the correct stored knowledge. While the empirical evidence in this paper is limited to the theorem-recognition setting, this issue arises in mathematical reasoning, scientific modeling, retrieval, verification, and structured decision making, where a known result or object may be expressed as a formula, constraint, optimization condition, set relation, invariant, rule, or intermediate characterization\cite{aamodt1994casebased,scharpf2023formula,alemi2016deepmath,yang2023leandojo}. Although these forms can preserve the same meaning, they may provide very different cues for recognition. A system may therefore appear to know a concept in its standard form while failing to identify or use it when the representation changes. This paper studies that gap between possessing knowledge and accessing it under equivalent transformation.

This problem is especially visible in mathematics, where the same content can be expressed through many equivalent forms. A theorem may appear as an inequality, an operator equality, a zero-residual condition, an existential witness statement, an optimization identity, a set-membership relation, or a proof-intermediate invariant. These forms can be equivalent while providing very different cues for recognition. Thus, theorem recognition offers a clean setting for studying whether a model can access known knowledge after the representation changes.

The ability to recover known results from transformed representations matters for more than benchmark accuracy. In formal theorem proving, a system often needs to select relevant premises before proof search can proceed; premise selection has long been identified as a major bottleneck in large mathematical libraries~\cite{alemi2016deepmath,yang2023leandojo}. Related work in mathematical information retrieval also shows that formula meaning cannot be reduced to surface notation: formula-concept recognition asks whether a formula can be matched to a unique underlying mathematical concept identifier~\cite{scharpf2023formula}. These settings all require access to formal knowledge under representational variation.

Current mathematical language-model evaluations do not directly isolate this ability. Benchmarks such as GSM8K, MATH, and Minerva-style evaluations test whether models can solve word problems, competition problems, or quantitative reasoning tasks and are usually scored by final answers or generated derivations \cite{cobbe2021training,hendrycks2021math,lewkowycz2022solving}. Formal-theorem-proving benchmarks evaluate proof construction or formalization ability \cite{polu2020generative,zheng2021minif2f,wu2022autoformalization}. Robustness benchmarks show that mathematical performance can degrade under perturbed or equivalent problem variants \cite{huang2025mathperturb,hao2025putnamgap}. However, these settings leave a complementary question underexplored: when the target is a known named theorem, can a model recover that theorem from an equivalent but unfamiliar formula-level representation?

The central question of this paper is whether formal knowledge remains accessible when its representation changes. We study this question through theorem recognition, asking whether models that report familiarity with a theorem in its standard form can recover it from an equivalent but unfamiliar mathematical representation. We ask four broad questions. First, to what extent do language models retain access to known formal knowledge when its representation changes? Second, when access fails, do models abstain, retrieve the wrong object, or produce unusable outputs? Third, how do exact accuracy and sample support vary across transformation buckets? Finally, how can test-time guidance improve this?

To address these questions, we build \textsc{TREAT}, a benchmark for Theorem Recognition under Equivalence-preserving mAthematical Transformation, using theorem recognition as a controlled testbed for studying representation-robust access to formal knowledge. Each item in the benchmark presents a transformed variant of a mathematical theorem. Unlike natural-language paraphrase benchmarks, the transformation is applied to the mathematical representation itself: theorem conditions may be rewritten as residual equations, witness statements, optimization identities, set relations, operator forms, distributional characterizations, or proof-intermediate forms.

The benchmark is constructed from theorems with mathematical expression forms. For each theorem, we extract a canonical theorem condition, record assumptions, and generate equivalence-preserving variants with inverse-mapping notes. 
Candidate variants are semantically validated with Gemini 3.1 Pro, transformations were encoded symbolically and checked with Z3, and each row records validation metadata.
The resulting corpus contains 737 theorem identities and 29,480 transformed rows, with metadata supporting validation, filtering, and error analysis.
We evaluate six language models on a 960-item shared-known panel, where the target theorem concepts are selected from theorems that multiple models report knowing in standard form. This design reduces the confound that a failure is simply due to unfamiliarity with the theorem. The best model achieves 60.73\% accuracy, with different systems having different patterns of abstention and incorrect theorem detection.

This paper makes four contributions. First, we formulate theorem recognition under equivalence-preserving transformation as a controlled evaluation of representation-robust access to formal knowledge. Second, we introduce \textsc{TREAT}, a benchmark with formula-level transformations, recorded assumptions, and validation metadata. Third, we provide a six-model evaluation showing that theorem recognition remains far from saturated even on frontier models. Fourth, we analyze failure behavior and test-time recovery, showing that missed theorem access can sometimes be recovered with additional guidance but must consider false-route risk.

\section{Related Work}

\subsection{Mathematical reasoning benchmarks.}
Most benchmarks for mathematical language models evaluate problem solving rather than recognition of known formal objects. GSM8K tests multi-step grade-school word problems \cite{cobbe2021training}, MATH evaluates competition-style problem solving \cite{hendrycks2021math}, and Minerva-style evaluations study quantitative reasoning over mathematical and scientific questions \cite{lewkowycz2022solving}. Formal mathematics benchmarks and systems, including GPT-f, MiniF2F, and autoformalization work, evaluate proof generation, formalization, or interaction with proof assistants \cite{polu2020generative,zheng2021minif2f,wu2022autoformalization}. These settings measure whether models can solve problems or construct proofs. In contrast, \textsc{TREAT} asks whether a model can recover the theorem identity associated with a transformed theorem condition. A model may know a theorem or solve related problems while still failing to identify the theorem after an equivalence-preserving change in representation.

\subsection{Robustness under mathematical variation.}
Recent work has shown that mathematical reasoning can be sensitive to changes in problem form. MathCheck evaluates mathematical reasoning using checklist-style variants \cite{zhou2024mathcheck}, GSM-Symbolic shows that GSM8K-style performance can degrade under symbolic template changes and irrelevant clauses \cite{mirzadeh2024gsmsymbolic}, MATH-Perturb constructs perturbed versions of difficult MATH problems \cite{huang2025mathperturb}, and PutnamGAP studies robustness under mathematically equivalent transformations of advanced problems \cite{hao2025putnamgap}. Related robustness concerns also arise in step-level mathematical verification, where evaluator judgments can change across semantically valid perturbations of solution traces \cite{mazdarani2026beyond}. TREAT is related to this line of work but differs in the target of evaluation. Prior robustness benchmarks typically ask whether a model can still solve a problem after perturbation. TREAT asks whether a model can retrieve the known theorem handle behind an equivalent but structurally unfamiliar formula-level representation.

\subsection{Formal knowledge retrieval and theorem access.}
TREAT also connects to work on retrieving reusable knowledge objects. Case-based reasoning studies retrieval and adaptation of prior cases \cite{aamodt1994casebased}; semantic parsing and entity linking map flexible inputs to structured meanings or knowledge-base entries \cite{zettlemoyer2005learning,berant2013semantic,shen2015entitylinking}. In mathematics, formula-concept recognition studies whether formulas can be matched to underlying mathematical concept identifiers despite variation in notation \cite{scharpf2023formula}. Premise selection and retrieval-augmented theorem proving show that formal proof systems often depend on identifying relevant prior results before proof search can proceed \cite{alemi2016deepmath,yang2023leandojo}. We isolate a complementary problem: given a transformed mathematical condition, can a language model recover the named theorem that indexes the relevant formal knowledge?

\section{Benchmark Methodology}
\begin{figure*}[t]
    \centering
    \includegraphics[width=\textwidth]{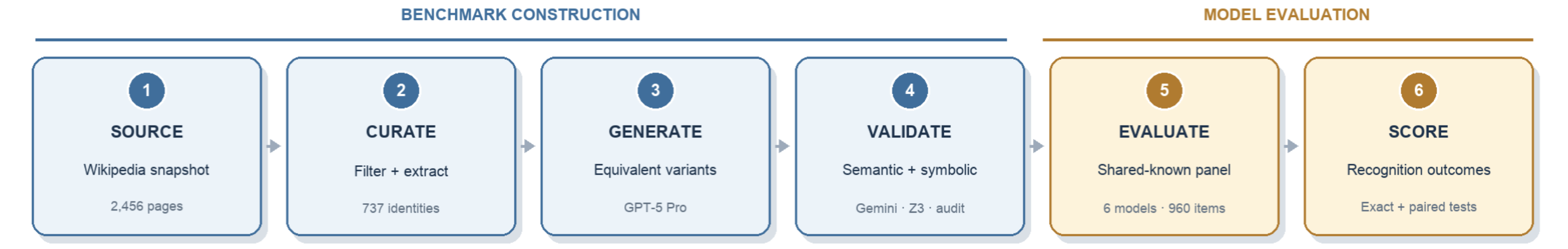}
    \caption{\textsc{TREAT} benchmark construction and evaluation workflow.}
    \label{fig:pipeline}
\end{figure*}
\subsection{Task Definition and Scope}

We study theorem recognition under equivalence-preserving transformation. Let $T$ denote a named theorem identity, $C_T$ its canonical mathematical condition, and $A_T$ the assumptions under which the theorem is stated. A transformed variant $V_{T,j}$ is intended to preserve the theorem condition under the same assumptions, i.e., to satisfy $A_T \models C_T \leftrightarrow V_{T,j}$. This biconditional is a target rather than an assumed property of the raw generated variants.
At evaluation time, the model receives only $V_{T,j}$, together with minimal surrounding text needed to parse the mathematical statement. It is not given the theorem name, theorem family, subfield, candidate list, canonical condition, or transformation type. The model must recover the theorem identity $T$, which serves as the handle for the canonical theorem statement stored in the benchmark.

This task differs from mathematical problem solving and proof generation. The goal is not to derive a numerical answer or construct a complete proof, but to determine whether a known theorem remains recognizable after its mathematical representation changes. Theorem recognition is a useful testbed because theorem identities provide stable targets, many theorem statements contain compact formula-level conditions, and equivalence can be audited through assumptions, inverse mappings, symbolic templates, and validation metadata.

\subsection{Benchmark Construction}

\subsubsection{Theorem Corpus}
\textsc{TREAT} begins with English Wikipedia to assemble a broad inventory of named mathematical results. Because pages may be incomplete, inconsistently structured, or later revised, we treat Wikipedia as a source inventory rather than mathematical ground truth. We retain named results with recoverable theorem conditions and assumptions; exclude broad topics, ambiguous objects, prose-only entries, duplicates, and non-condition formulas; and then apply the validation below. Redirects and near-duplicates are consolidated under a canonical label. Each of the 737 retained identities stores its name, aliases, field, canonical statement and condition $C_T$, and assumptions $A_T$. For reproducibility, the underlying 2,456-page crawl was retained on March 27, 2026. The exact archived text, titles, URLs, categories, and checksum, together with the benchmark, schemas, prompts, model-provenance record, exact-correctness vectors, and paired-test code, are available in the public \textsc{TREAT} repository~\cite{treatrepo2026}.
\subsubsection{Variant Generation}

\newcolumntype{Y}{>{\raggedright\arraybackslash}X}

\begin{table*}[t]
\caption{Transformation types used to disguise a theorem concept while preserving its canonical formal content. These are formula-level transformations, not natural-language paraphrases.}
\label{tab:transforms}
\centering
\scriptsize
\setlength{\tabcolsep}{3pt}
\renewcommand{\arraystretch}{1.15}
\begin{tabularx}{\textwidth}{>{\raggedright\arraybackslash}p{0.17\textwidth}
>{\raggedright\arraybackslash}p{0.30\textwidth}
>{\raggedright\arraybackslash}p{0.21\textwidth}
Y}
\toprule
Type & Equivalence schema & Example form & Retrieval challenge \\
\midrule
Residual / nonnegativity & $F=0 \Leftrightarrow \|F\|^2=0$; $a\le b \Leftrightarrow (a-b)_+=0$ & Replace equality or inequality by zero-loss form & Hides the concept as a residual condition. \\
Slack / witness & $a\le b \Leftrightarrow \exists s\ge0: b=a+s$ & Introduce an existential variable & Converts an order statement into a witness statement. \\
Monotone embedding & $a\le b \Leftrightarrow \phi(a)\le\phi(b)$ for strictly increasing $\phi$ & Exponential, log, or sigmoid-style order encoding & Preserves truth while changing scale and surface form. \\
Optimization / projection & $a\le b \Leftrightarrow \min_{s\ge0}(b-a-s)^2=0$ & Zero optimum or projection distance & Encodes a theorem condition as an extremal identity. \\
Set membership / equality & $x=y \Leftrightarrow (x,y)\in\Delta$; $S\subseteq T \Leftrightarrow S\cap T^c=\varnothing$ & Diagonal, containment, cylinder, or fiber encoding & Recasts algebraic or logical equality as set structure. \\
Quantifier / logic & $\exists x P(x) \Leftrightarrow \neg\forall x\neg P(x)$ & Negation-normal or truth-table form & Changes the quantifier surface while preserving logical content. \\
Operator / functional & $f=g \Leftrightarrow \|f-g\|=0$; $Tx=Sx$ for all $x$ & Norm, kernel, or operator equality & Recasts function-level statements as operator conditions. \\
Integral transform & $f=g \Leftrightarrow \mathcal{T}f=\mathcal{T}g$ when $\mathcal{T}$ is injective & Fourier, Laplace, or moment-style encoding & Represents equality or distributional identity through a transform. \\
Counting / combinatorial & $|S|=n \Leftrightarrow \sum_{x\in S}1=n$ & Indicator-sum or incidence form & Rewrites discrete claims as algebraic counting identities. \\
Probabilistic / distributional & $X\sim\mu \Leftrightarrow \mathbb{E}f(X)=\int f\,d\mu$ for a determining class & Expectation, tail, or distributional characterization & Changes a probability theorem into an equivalent distributional condition. \\
Proof-intermediate / characterization & $C_K \Leftrightarrow I_K$ for a theorem-specific intermediate $I_K$ & A known equivalent lemma, invariant, or iff condition & Tests retrieval through a nonstandard but exact characterization. \\
\bottomrule
\end{tabularx}
\end{table*}

The benchmark is designed to intervene on mathematical representation rather than on surface wording. We define a taxonomy of equivalence-preserving transformation families that include slack-witness encodings of inequalities, residual and nonnegativity encodings, singleton and diagonal equality encodings, set-membership reformulations, optimization and projection identities, monotone order embeddings, operator and functional encodings, integral transform characterizations, probabilistic and distributional reformulations, counting encodings, and theorem-specific proof-intermediate characterizations. Table~\ref{tab:transforms} reports representative schemas for these transformation families.
The examples are schematic because the concrete instantiation depends on the theorem's domain, variables, and assumptions. 

Given the retained theorem inventory and the predefined transformation taxonomy, GPT-5 Pro is used to generate candidate variants. Each prompt is conditioned on the theorem name, canonical statement, assumptions, canonical condition, and a target transformation family. The model is asked to produce a transformed mathematical condition together with an inverse-mapping note explaining how the transformed form reduces back to the canonical condition. This makes generation theorem-conditioned and transformation-conditioned rather than open-ended. 
The final benchmark consists of 29,480 transformed rows with a nonuniform number of variants per theorem. It is an auditable collection of transformable theorem identities, not a census of mathematical theorems or transformations. The pipeline is: filter pages, extract conditions, generate variants, validate them semantically and symbolically, inspect a subset manually, then evaluate and score models.
Fig.~\ref{fig:pipeline} summarizes the workflow; Appendix~\ref{app:prompts} records the prompt contracts.

\subsubsection{Validation}
After variant generation, Gemini 3.1 Pro judges each candidate as equivalent, equivalent under assumptions, wrong, stronger, weaker, or unclear from its theorem metadata, formulas, transformation type, and inverse mapping. Wrong rows are removed; stronger, weaker, or unclear rows are rejected or reviewed further and excluded. Separating GPT-5 Pro generation from Gemini validation avoids direct self-approval, but changing either model could change the corpus. Both model families also appear in the evaluation, so construction and evaluation are not fully independent.

Z3 was used as another validation layer. While it does not prove the source theorem itself, it checks whether the transformation template preserves the encoded theorem condition\cite{demoura2008z3}. For a canonical condition $C$ and transformed condition $V$, the checker asserts the negated equivalence,
\begin{equation}
(C \wedge \neg V) \vee (\neg C \wedge V),
\end{equation}
and accepts the template-level check when this formula is unsatisfiable under the recorded abstraction and assumptions.

For example, a slack transformation for an inequality checks that $a\le b$ is equivalent to $\exists s\ge0:b=a+s$ over the relevant numeric domain. A residual transformation checks that $F=0$ is equivalent to $F^2=0$ or $|F|^2=0$ under nonnegativity assumptions. Analytic transformations, distributional characterizations, and theorem-specific proof-intermediates are not always directly decidable in Z3; those rows retain semantic and metadata validation, and the Z3 layer is recorded as template-level, axiomatized, or not applicable. The checks pass successfully and do not find any counterexample among the symbolically encodable cases.

In addition, we manually inspect a 100-row validation subset consisting of 10 theorem identities with 10 transformed variants per theorem selected randomly. We did not identify any mathematically wrong mapping or counterexample. 

\section{Evaluation Methodology}

\subsection{Evaluation Panel Selection}

Before transformed evaluation, each model receives the standard-form familiarity probe in Appendix~\ref{app:prompts}, which requires a known/unknown judgment and brief standard content. We sample 160 identities from the intersection reported known by all models and six variants per identity, yielding 960 items. This canonical-form familiarity control reduces complete unfamiliarity, although it is not an independently scored matched canonical-condition task. We selected the six models to span commercial and open-weight access, general and coding-oriented training, and several scales, testing whether representation access generalizes across deployment regimes rather than one model family. Names are provider labels at access time.

The 960-item panel covers five broad mathematical families: 504 Analysis items, 180 Algebra/Linear Algebra items, 156 Probability/Information Theory items, 78 Number Theory items, and 42 Combinatorics/Graph Theory items. It also spans nine transformation buckets. The transformation-bucket composition is reported in Fig.~\ref{fig:transformdist}. Buckets with small support, such as residual/nonnegativity and characterization/iff, are included for coverage but are not used for strong bucket-level conclusions. Family labels are assigned at the theorem-identity level using the source page categories, field metadata, and the primary mathematical context of the canonical statement. Boundary cases are assigned to the family most directly associated with the theorem identity. We therefore treat family accuracy as a coarse diagnostic of neighborhood-level retrieval, not as a complete mathematical ontology.

\subsection{Recognition Task}

Each evaluation input contains only the transformed mathematical statement and the minimal surrounding text needed to parse it. The model is not given the theorem family, subfield, candidate theorem list, original theorem name, canonical condition, or transformation type. It must return a structured JSON response indicating whether a valid theorem route exists, the predicted theorem name or \none, relevant aliases, a short connection explaining the match, a brief justification sketch, any missing assumptions, and a confidence score.
The theorem name is used as the scored handle for the canonical theorem form stored in the benchmark. This is a practical choice because named theorem identities have aliases, field labels, and canonical statements that can be normalized for evaluation. The explanatory fields are not the primary target of scoring; they are included to discourage unsupported name guessing and to support later audit of model behavior.

\subsection{Scoring Metrics}

The primary metric is exact theorem-identification accuracy. A prediction is correct when the normalized name matches the gold identity. Normalization handles capitalization, punctuation, possessives, parenthetical aliases, selected synonyms, and minor eponym-order variations. Family accuracy credits predictions identifying a theorem from the correct mathematical family. Small exact-to-family gaps suggest that minor naming variants do not drive the headline result. Paired exact-accuracy comparisons use Cochran's $Q$, two-sided exact McNemar tests with Holm correction over 15 pairs, and a theorem-cluster sign-flip sensitivity analysis.

To characterize model behavior beyond top-line accuracy, we report four additional quantities. The asserted rate is the share of items for which the model returns a theorem name rather than \none. The wrong-named rate measures cases where the model asserts a theorem identity but the identity is incorrect. The \none{} rate measures abstention or failure to retrieve a theorem concept. The malformed-output rate measures invalid JSON or responses that violate the required schema. These metrics are important because models with similar exact accuracy may behave very differently: one may aggressively commit to theorem names and incur more wrong-theorem errors, while another may abstain on many recoverable items.

\subsection{Test-Time Computation Setting}

As a secondary analysis, test-time computation (TTC) tests whether retry prompts, family or subfield hints, or transformation-type hints recover missed recognitions; it does not define the primary score. We apply each intervention to valid rows and matched negative or bait rows whose answer is \none. An intervention is useful only if it improves valid-item recognition without substantially increasing false theorem routes.

\section{Results}

The panel contains 960 transformed variants of 160 theorem concepts, with six variants per concept. Every item has a gold canonical theorem handle by construction, so a \none{} response is a failed retrieval on this benchmark rather than a true negative.

\begin{table}[t]
\caption{Shared-known 960-item canonical-form retrieval results.}
\label{tab:leaderboard}
\centering
\scriptsize
\resizebox{\columnwidth}{!}{%
\begin{tabular}{lrrrrrrrr}
\toprule
Model & Asserted & Exact & Family & Wrong named & \none & Malformed \\
\midrule
GPT-5.4 Pro & 82.19\% & 60.73\% & 64.38\% & 17.81\% & 17.81\% & 0.00\% \\
Gemma 3 12B & 61.98\% & 52.81\% & 53.85\% & 8.12\% & 38.02\% & 0.00\% \\
Gemini 3.1 Pro & 70.31\% & 52.50\% & 53.75\% & 16.56\% & 29.69\% & 0.00\% \\
Gemma 3 27B & 53.02\% & 46.98\% & 47.81\% & 5.21\% & 46.98\% & 1.67\% \\
Qwen3 Coder 480B & 50.00\% & 46.88\% & 47.50\% & 2.50\% & 50.00\% & 25.21\% \\
GPT-OSS 120B & 13.44\% & 11.15\% & 11.46\% & 1.98\% & 86.56\% & 0.00\% \\
\bottomrule
\end{tabular}}
\end{table}

\subsection{Access Under Representation Change}

Table~\ref{tab:leaderboard} gives the main answer to the first question. Recognition remains far from saturated even under the shared-known design. GPT-5.4 Pro is the strongest system, with 583 exact matches out of 960 items, or 60.73\%, but it still fails to retrieve the correct theorem identity on 377 transformed variants. The middle group ranges from 46.88\% to 52.81\%, while GPT-OSS 120B reaches 11.15\%. Thus, under this self-reported familiarity control, access remains limited after an equivalence-preserving representation change.

Cochran's $Q$ rejects equal exact accuracy across models ($Q(5)=740.75$, $p=7.57\times10^{-158}$). Holm-adjusted exact McNemar tests identify two ties: Gemma 3 12B versus Gemini 3.1 Pro and Gemma 3 27B versus Qwen3 Coder 480B (both adjusted $p=1.00$); all other item-level pairs differ ($p\leq0.0245$). The theorem-cluster sensitivity retains GPT-5.4 Pro over Gemma 3 27B, Qwen3, and GPT-OSS, and every non-GPT-OSS model over GPT-OSS (adjusted $p\leq0.0014$); other contrasts are not retained. Full results and code are in the repository.

The small gap between exact accuracy and family accuracy is also informative. For most models, family accuracy improves exact accuracy by only about 0.6--3.7 percentage points. This means that many failures are not merely alias or near-miss problems within the right area of mathematics. They often reflect either abstention or retrieval of a different theorem concept. In this retrieval setting, the model does not just need to know the broad domain; it must recover the correct formal handle.

\subsection{Failure Behavior: Abstention, Wrong Retrieval, and Output Reliability}

Top-line accuracy hides different access policies. Table~\ref{tab:policy} conditions on the cases where a model asserted a theorem concept rather than returning \none. GPT-5.4 Pro has the highest coverage, asserting a theorem on 789 items, but only 73.9\% of those assertions are exact and 21.7\% are wrong named-theorem commitments. Gemini 3.1 Pro shows a similar high-coverage profile, with 675 asserted theorem names, 74.7\% exact among assertions, and 23.6\% wrong among assertions. By contrast, Qwen3 Coder 480B asserts on only 480 items, but 93.8\% of those assertions are exact and only 5.0\% are wrong. Gemma 3 27B is similarly selective, with 509 assertions, 88.6\% exact among assertions, and 9.8\% wrong among assertions.

\begin{table}[t]
\caption{Retrieval-policy decomposition. Asserted $n$ is the number of items on which the model returned a theorem concept. Conditional exact and wrong rates are computed over asserted items. Malformed output is reported separately.}
\label{tab:policy}
\centering
\scriptsize
\begin{tabular}{lrrrrr}
\toprule
Model & $n$ & Exact & Wrong & Abstained $n$ & Malformed \\
\midrule
GPT-5.4 Pro & 789 & 73.9\% & 21.7\% & 171 & 0.00\% \\
Gemma 3 12B & 595 & 85.2\% & 13.1\% & 365 & 0.00\% \\
Gemini 3.1 Pro & 675 & 74.7\% & 23.6\% & 285 & 0.00\% \\
Gemma 3 27B & 509 & 88.6\% & 9.8\% & 451 & 1.67\% \\
Qwen3 Coder 480B & 480 & 93.8\% & 5.0\% & 480 & 25.21\% \\
GPT-OSS 120B & 129 & 82.9\% & 14.7\% & 831 & 0.00\% \\
\bottomrule
\end{tabular}
\end{table}

\begin{figure}[t]
\centering
\includegraphics[width=\linewidth]{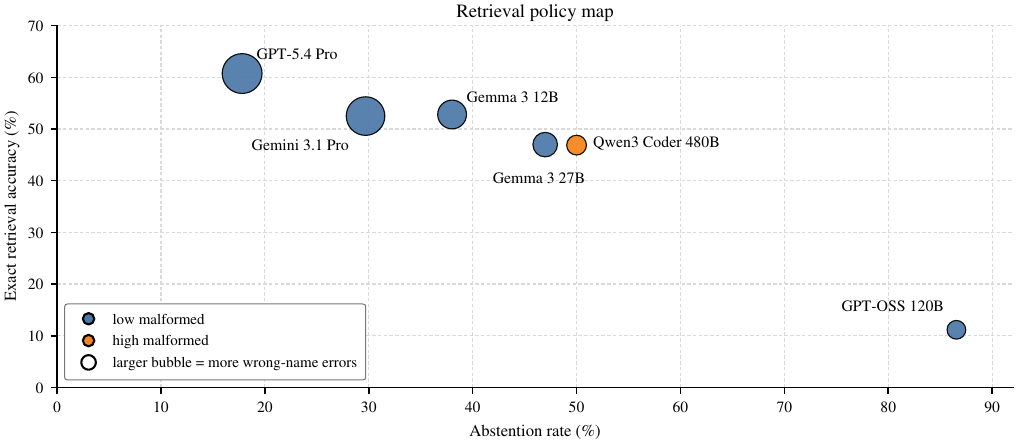}
\caption{Retrieval-policy map on the 960-item panel. The x-axis shows abstention/\none{} rate, the y-axis shows exact retrieval accuracy, and bubble size reflects wrong named-theorem assertions.}
\label{fig:profiles}
\end{figure}

Fig.~\ref{fig:profiles} visualizes these differences. The models are not arranged along a single ability axis. GPT-5.4 Pro and Gemini 3.1 Pro retrieve often, which gives them higher all-item recall but also a larger wrong-theorem risk. Qwen3 Coder 480B and Gemma 3 27B are more conservative: when they commit, they are usually correct, but they leave roughly half of the panel unresolved. GPT-OSS 120B is dominated by abstention, returning \none{} on 831 items.

This distinction matters because the failure modes have different consequences. A wrong theorem route may send a proof search, verifier, retrieval system, or explanation toward the wrong formal object. Abstention is safer but limits usefulness when a system needs to hand off to a formal artifact. Structured-output reliability adds a third dimension to the error analysis. The malformed-output rates are reported in Tables~\ref{tab:leaderboard} and~\ref{tab:policy}; Qwen3 Coder 480B's malformed outputs persisted across repeated runs under the same structured-output protocol. It recorded a 25.21\% malformed-output rate, despite having the highest conditional precision among asserted theorem names in Table~\ref{tab:policy}. This separates mathematical recoverability from interface reliability: a model may often identify the right theorem when it commits, yet still be difficult to use in a structured task. 

\subsection{Where Representation Change Is Easier or Harder}

\begin{figure}[t]
\centering
\includegraphics[width=\linewidth]{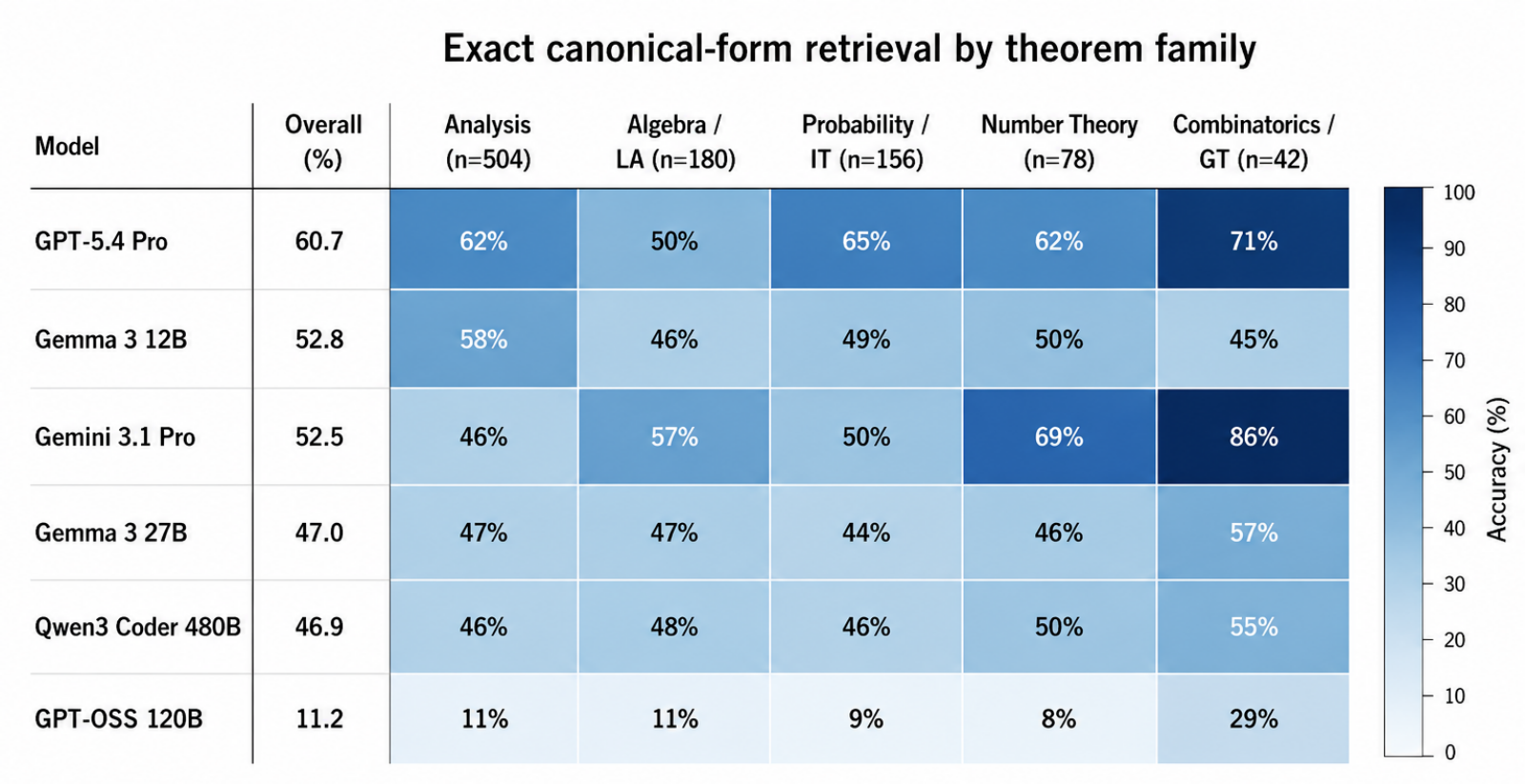}
\caption{Family-level exact retrieval accuracy heatmap. Numbers are percentages. Small-family cells, especially Combinatorics/Graph Theory, should be read as diagnostic rather than definitive.}
\label{fig:familyheat}
\end{figure}

We next ask which properties of the transformed representation affect recognition. Fig.~\ref{fig:familyheat} visualizes exact theorem identification by mathematical family matrix. Analysis family has the largest support at 504 items, so its estimates are the most stable. Combinatorics/Graph Theory has only 42 items, so results in that column should be treated as diagnostic rather than definitive.

The family matrix argues against treating theorem recognition as a single undifferentiated math-knowledge score. GPT-5.4 Pro is the strongest aggregate system, but Gemini 3.1 Pro is higher on Algebra/Linear Algebra, Number Theory, and the small Combinatorics/Graph Theory slice. GPT-5.4 Pro is stronger on Analysis and Probability/Information Theory. These differences suggest that recognition depends on the interaction between the transformed representation and the mathematical neighborhood in which the theorem lives. Some families contain distinctive formal signatures, while others contain broad inequalities, extremal principles, or analytic conditions that overlap across multiple named results.

\begin{figure}[t]
\centering
\includegraphics[width=\linewidth]{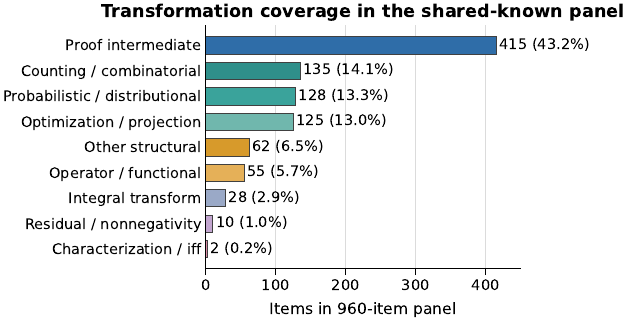}
\caption{Transformation-bucket distribution of the 960-item panel.}
\label{fig:transformdist}
\end{figure}

Transformation type is another source of variation. Fig.~\ref{fig:transformdist} reports the transformation-bucket distribution in the shared-known panel, while Table~\ref{tab:buc} ranks buckets by six-model mean exact accuracy. Optimization/projection variants are the hardest major bucket, with a six-model mean of 37.9\%. Integral transform variants have the highest mean at 58.3\%, but their support is only 28 items, so they should not be overinterpreted. Among larger buckets, counting/combinatorial, proof-intermediate, and probabilistic/distributional transformations cluster around 45\% mean exact accuracy. These results show that the benchmark is not merely testing theorem rarity. A theorem can become easier or harder to recognize depending on whether it is exposed through an optimization identity, a counting invariant, a distributional condition, or a proof-intermediate characterization.

\begin{table}[t]
\caption{Transformation buckets with at least 10 items, ranked by six-model mean exact accuracy.}
\label{tab:buc}
\centering
\scriptsize
\begin{tabular}{lrr}
\toprule
Bucket & $n$ & Mean exact \\
\midrule
Integral transform & 28 & 58.3\% \\
Other structural & 62 & 53.0\% \\
Operator / functional & 55 & 50.3\% \\
Counting / combinatorial & 135 & 45.2\% \\
Proof intermediate & 415 & 44.9\% \\
Probabilistic / distributional & 128 & 44.7\% \\
Residual / nonnegativity & 10 & 43.3\% \\
Optimization / projection & 125 & 37.9\% \\
\bottomrule
\end{tabular}
\end{table}

\subsection{Test-Time Recovery and False-Route Risk}

The final question is whether missed recognitions can be recovered without encouraging unsafe over-recognition. The main results above use the zero-hint setting. We therefore evaluate test-time computation (TTC) as a secondary analysis. TTC means additional second-pass computation or additional prompt-side information; model parameters are not updated. On positive rows, TTC is applied only when the baseline row is a detectable failure, either because the model returned a false route decision or because the output was malformed. Baseline exact successes are preserved. On negative or bait rows, each row is forced through the same TTC method using a fake failed baseline. These negative rows are paired with tempting source theorems, but their displayed statements are invalid or perturbed, so the correct answer is \texttt{NONE}.

We evaluate retry-after-failure prompting, family and subfield hints, and transformation-type hints. Table~\ref{tab:ttc} reports the main TTC results. The strongest recovery pattern appears for Qwen3. Retry, family hints, and transform hints all improve exact accuracy by more than 26 percentage points while producing 0/960 false routes on matched negative controls. Family hinting is the best method for Qwen, improving from 46.88\% to 77.50\%, a gain of 30.62 points. 
The TTC results also show that positive recovery and safety are different properties. Gemma 27B illustrates the risk most clearly: family hints improve positive accuracy by 16.15 points, but produce 329 false named routes out of 960 negative rows. Gemma 12B family hints are safer in this sweep, improving by 17.29 with 0/960 false routes. Thus, the same intervention type can be helpful for one model and unsafe for another.

\begin{table}[t]
\caption{Test-time computation on 960 positive and 960 matched
negative/bait items. Baseline and final report exact accuracy (\%);
gain is in percentage points (pp); false routes are named predictions
on negatives (count/960).}

\label{tab:ttc}
\centering
\scriptsize
\setlength{\tabcolsep}{2.5pt}
\renewcommand{\arraystretch}{1.05}
\resizebox{\columnwidth}{!}{%
\begin{tabular}{@{}llrrrc@{}}
\toprule
Model & Method &
\shortstack{Baseline(\%)} &
\shortstack{Final (\%)} &
\shortstack{Gain(pp)} &
\shortstack{False routes} \\
\midrule
Qwen3 Coder 480B & Retry if fail  & 46.88 & 74.90 & +28.02 & 0/960   \\
                  & Family hint    & 46.88 & 77.50 & +30.62 & 0/960   \\
                  & Transform hint & 46.88 & 73.65 & +26.77 & 0/960   \\
\addlinespace[2pt]
Gemma 3 12B      & Retry if fail  & 52.81 & 60.83 & +8.02  & 93/960  \\
                  & Family hint    & 52.81 & 70.10 & +17.29 & 0/960   \\
                  & Transform hint & 52.81 & 59.48 & +6.67  & 117/960 \\
\addlinespace[2pt]
Gemma 3 27B      & Retry if fail  & 46.98 & 58.33 & +11.35 & 185/960 \\
                  & Family hint    & 46.98 & 63.12 & +16.15 & 329/960 \\
                  & Transform hint & 46.98 & 60.73 & +13.75 & 88/960  \\
\bottomrule
\end{tabular}%
}
\end{table}

\section{Discussion}

\subsection{Accessibility Under Representation Change}

For theorem concepts with usable mathematical forms, models familiar with the standard form often fail to recover the identity from an equivalent variant. The shared-known design makes complete absence of the theorem less plausible, so these failures more plausibly concern access under representation change, although the design does not reveal models' internal representations.

Formal knowledge is useful only when connected to the forms encountered in practice. In \textsc{TREAT}, a named theorem is an auditable handle for a stored canonical statement and condition. In other domains, the target might be a library entry, verification rule, schema element, ontology identifier, invariant, or executable specification. We do not claim that theorem-domain accuracies numerically predict those settings. Rather, the theorem task tests a shared requirement: equivalent but unfamiliar representations should still retrieve the formal object needed for use or audit.

\subsection{Accuracy, Abstention, and Wrong Retrieval}

Top-line exact accuracy is only one part of representation-robust access. Models differ not only in how often they recover the correct theorem, but also in how they fail. Some failures are abstentions: the model returns \texttt{NONE} even though every main benchmark item has a gold theorem identity. GPT-OSS 120B is the clearest example, with 831 abstentions out of 960 items, while Qwen3 Coder 480B and Gemma 3 27B abstain on roughly half of the panel. Other failures are wrong commitments: the model asserts a theorem name, but the theorem is incorrect. GPT-5.4 Pro and Gemini 3.1 Pro have higher coverage, but roughly one-fifth to one-quarter of their asserted theorem names are wrong. A third failure mode is interface failure, where the model does not satisfy the required structured-output format, as reflected in Qwen3 Coder 480B 's 25.21\% malformed-output rate.

These errors have different consequences. A wrong theorem route may direct a proof search, verifier, retrieval system, or explanation toward the wrong formal object. Abstention is safer, but it limits usefulness. The policy decomposition therefore changes how the leaderboard should be interpreted. Qwen3 Coder 480B has lower all-item exact accuracy than Gemma 3 12B, but among asserted theorem names it is exact on 93.8\% of cases. GPT-5.4 Pro obtains the highest all-item exact score because it covers more of the panel, but this comes with more wrong-theorem risk. For formal applications, the preferred operating point may depend on whether the system values coverage, precision, abstention, or schema reliability.

\subsection{Structured Representation Effects}

The family and transformation analyses show that representation failure is not uniform. Recognition varies across mathematical families, transformation types, and theorem identities. This argues against treating theorem recognition as a single undifferentiated measure of mathematical knowledge. A theorem with a distinctive algebraic signature may remain recognizable after some transformations, while a theorem expressed through a generic inequality, extremal principle, convergence statement, or distributional condition may be easier to confuse with nearby results.

The transformation results are especially important for interpreting what the benchmark measures. \textsc{TREAT} does not merely paraphrase theorem text. It changes the mathematical form of the theorem condition through witness encodings, residual forms, optimization identities, set-membership reformulations, operator forms, distributional characterizations, and proof-intermediate invariants. These transformations can preserve truth under the recorded assumptions while changing the route by which the theorem is recognized. This places the benchmark between ordinary name recognition and full proof reconstruction: the model is not asked to prove the theorem from first principles, but it must recover the correct stored theorem handle when familiar surface cues are removed.

\subsection{Test-Time Guidance}

The TTC results add a second layer to the main finding. Some failures are recoverable, which suggests that representation failure is not always an absence of theorem knowledge. In several cases, the model appears to need help localizing the relevant theorem neighborhood. The results indicate that narrowing the search space to a coherent mathematical neighborhood can make latent theorem access more effective.

The TTC results also show that guidance must be evaluated for safety, not only for positive recovery. Matched negative and bait rows are essential because an intervention can improve recognition on valid theorem variants while also encouraging false theorem routes on invalid statements. Gemma 27B illustrates this risk: family hints improve positive accuracy by 16.15 points, but produce 329 false named routes out of 960 negative rows, or 34.27\%. By contrast, Gemma 12B family hints are safer in this sweep, giving a 17.29-point gain with no false routes on matched negatives. Thus, the same intervention type can be useful for one model and unsafe for another.

Overall, the TTC study changes the interpretation of the benchmark. The zero-hint results show that representation change disrupts direct access to known theorem identities. The TTC results show that some of this lost access can be recovered with additional localization. But the negative controls show that recovery must be coupled with refusal behavior: a useful intervention must improve recognition on valid transformed statements while preserving the ability to return \texttt{NONE} when no valid theorem route exists.

\section{Limitations and Future Work}
This study has several limitations. First, \textsc{TREAT} is restricted to theorem concepts with recoverable mathematical expression forms. Many mathematical results are conceptual, geometric, algorithmic, or prose-based, and are therefore outside the current benchmark. The benchmark should thus be read as measuring representation-robust theorem recognition for a subset of theorems, not for all mathematical knowledge.

Second, the corpus is shaped by its source inventory and filtering pipeline. The scraped theorem pages vary in coverage, notation, and quality, and some theorem families and transformation buckets are better represented than others. Small-support buckets should therefore be interpreted cautiously. Future versions should expand and better balance theorem families, subfields, and transformation types to separate domain effects from representation effects.

Third, although transformed variants are filtered using explicit assumptions, inverse mappings, and validation labels and Z3 template checks and a small human audit confirmed the validity, the corpus is still not proof-assistant-certified. Z3 applies only to encodable transformation templates, and analytic, distributional, and proof-intermediate variants may require assumptions or reasoning outside the SMT fragment. Another validation risk is stronger/weaker drift and missing side condition leakage. Therefore, larger expert audits and proof-assistant formalization for selected subsets are important directions for future work.

Fourth, the shared-known panel applies a standard-form familiarity control: every included theorem was reported known by all evaluated models, with brief standard content requested. This makes complete unfamiliarity less plausible, although the probe was not independently scored as a matched canonical-condition task. We cannot determine whether transformed formulations appeared in pretraining, a general difficulty for black-box models~\cite{oren2023contamination}. Similarly, the TTC study does not exhaust retrieval, verification, constrained decoding, tool use, or multi-step reasoning strategies.

Future work should study stronger recovery mechanisms such as theorem-library search, symbolic preprocessing, proof-state context, calibrated abstention, and constrained decoding. These methods should be evaluated with both positive transformed items and matched invalid controls, so that improved recovery is not confused with unsafe over-recognition. 

Finally, future work could test whether this evaluation protocol transfers beyond mathematics, to domains such as software verification, database query rewriting, scientific equation retrieval, ontology linking, or rule-based decision systems. Such extensions would require domain-specific target objects, equivalence relations, transformation families, and validation procedures.

\appendices
\section{Prompt Details}
\label{app:prompts}

For theorem recognition, the prompt contract was: ``You are given a transformed mathematical statement. The statement is intended to be mathematically equivalent to the canonical formal form of a known mathematical concept, usually a named theorem, but it may be written in a noncanonical formula or condition form. Task: retrieve the underlying mathematical concept and its standard theorem identity. Treat the theorem identity as the handle for the canonical formal form. Do not rely on topic metadata, theorem lists, family hints, or candidate names. Use only the mathematical content of the transformed statement. Return valid JSON only with fields \texttt{recognized}, \texttt{theorem\_name}, \texttt{aliases}, \texttt{connection}, \texttt{justification\_sketch}, \texttt{missing\_assumptions}, and \texttt{confidence}. If no named theorem concept can be responsibly retrieved, set \texttt{recognized} to \texttt{false} and \texttt{theorem\_name} to \texttt{NONE}.'' 

Before transformed-form evaluation, models were also given a familiarity probe: ``You are given a list of named mathematical theorems. For each theorem, answer whether you recognize the theorem in its standard form. Mark a theorem as known only if you can identify the standard theorem concept and would be able to state its usual mathematical content. Return JSON with \texttt{theorem\_name}, \texttt{known}, and \texttt{brief\_standard\_content}.'' 
The shared-known panel was sampled from theorem identities marked known by the familiarity-probed models, with six transformed variants per selected theorem.

\bibliographystyle{IEEEtran}
\bibliography{refrences}
\end{document}